\documentclass[10pt,twocolumn,letterpaper]{article}

\usepackage[pagenumbers]{iccv} %

\usepackage{epsfig}
\usepackage{listings}
\usepackage{amsmath}
\usepackage{amssymb}
\usepackage{url}
\usepackage{amsmath} %
\usepackage{amssymb} %
\usepackage{multirow}
\usepackage{makecell} %
\usepackage{bm} %
\usepackage{stmaryrd} %
\usepackage{dsfont} %
\usepackage{pifont} %
\usepackage{ifthen} %
\usepackage{booktabs}
\usepackage[dvipsnames]{xcolor}
\usepackage{caption}  %
\newcommand\blfootnote[1]{%
  \begingroup
  \renewcommand\thefootnote{}\footnote{#1}%
  \addtocounter{footnote}{-1}%
  \endgroup
}
\usepackage[accsupp]{axessibility}  %

\newcommand{\dataset}[1]{{\textsc{PlaceIt3D}}}
\newcommand{\network}[1]{{\textsc{PlaceWizard}}}

\newcommand{\R}{\mathbb{R}} %

\newcommand{\B}{\{0,1\}} %
\newcommand{\Bint}{[0,1]} %

\newcommand{\PC}{\mathbf{X}} %
\newcommand{\mask}{\mathcal{M}} %
\newcommand{\point}{\mathbf{x}} %

\newcommand{\Lo}{\mathcal{L}} %

\newcommand{\loc}{\mathrm{[LOC]}}
\newcommand{\seg}{\mathrm{[SEG]}}
\newcommand{\rot}{\mathrm{[ROT]}}
\newcommand{\anc}{\mathrm{[ANC]}}

\newcounter{ablationrow}
\renewcommand{\theablationrow}{\Alph{ablationrow}}
\newcommand{\row}[1]{\refstepcounter{ablationrow}\label{#1}\theablationrow}

\newcommand{\projecturl}{https://nianticlabs.github.io/placeit3d/}

\definecolor{keywordcolor}{RGB}{0, 0, 139}   %
\definecolor{backgroundcolor}{RGB}{245, 245, 245} %
\definecolor{numbercolor}{RGB}{0, 100, 0}      %
\definecolor{functioncolor}{RGB}{255, 99, 71}  %
\definecolor{acolor}{RGB}{104, 180, 160}
\definecolor{a1color}{RGB}{255, 69, 0}   
\definecolor{a2color}{RGB}{255, 192, 100}
\lstdefinelanguage{YAML}{
  keywords={relationships, name, templates, special_templates},
  keywordstyle=\color{keywordcolor}\bfseries,
  columns=fullflexible,
  sensitive=true,
  basicstyle=\ttfamily\footnotesize,
  showstringspaces=true,
  breaklines=true,
  morestring=[b]",
  stringstyle=\color{black},
  morecomment=[l]{\#},
  literate={anchor_class}{\bfseries{\textcolor{acolor}{\{anchor\_class\}}}}1
  {anchor1_class}{\bfseries{\textcolor{a1color}{\{anchor1\_class\} }}}1
  {anchor2_class}{\bfseries{\textcolor{a2color}{\{anchor2\_class\} }}}1
}

\definecolor{rOneColor}{rgb}{0.95, 0.52, 0.25}   %
\definecolor{rTwoColor}{rgb}{0.88, 0.45, 0.55}   %
\definecolor{rThreeColor}{rgb}{0.3, 0.6, 0.88}   %

\definecolor{iccvblue}{rgb}{0.21,0.49,0.74}
\usepackage[pagebackref,breaklinks,colorlinks,allcolors=iccvblue]{hyperref}

\title{\textsc{PlaceIt3D}: Language-Guided Object Placement in Real 3D Scenes}

\author{Ahmed Abdelreheem$^{2,*}$\quad Filippo Aleotti$^{1}$\quad Jamie Watson$^{1}$\quad Zawar Qureshi$^{1}$\quad Abdelrahman Eldesokey$^{2}$ \\ 
Peter Wonka$^{2}$\quad Gabriel Brostow$^{1,3}$\quad Sara Vicente$^{1}$\quad Guillermo Garcia-Hernando$^{1}$\\
$^{1}$Niantic Spatial \quad $^{2}$KAUST \quad $^{3}$UCL\\\\
 \url{https://nianticlabs.github.io/placeit3d/}
}

\begin{document}

\twocolumn[{
\maketitle
\begin{center}
\captionsetup{type=figure}
\centering
\vspace{-1cm}
\includegraphics[scale=1,width=1\linewidth]{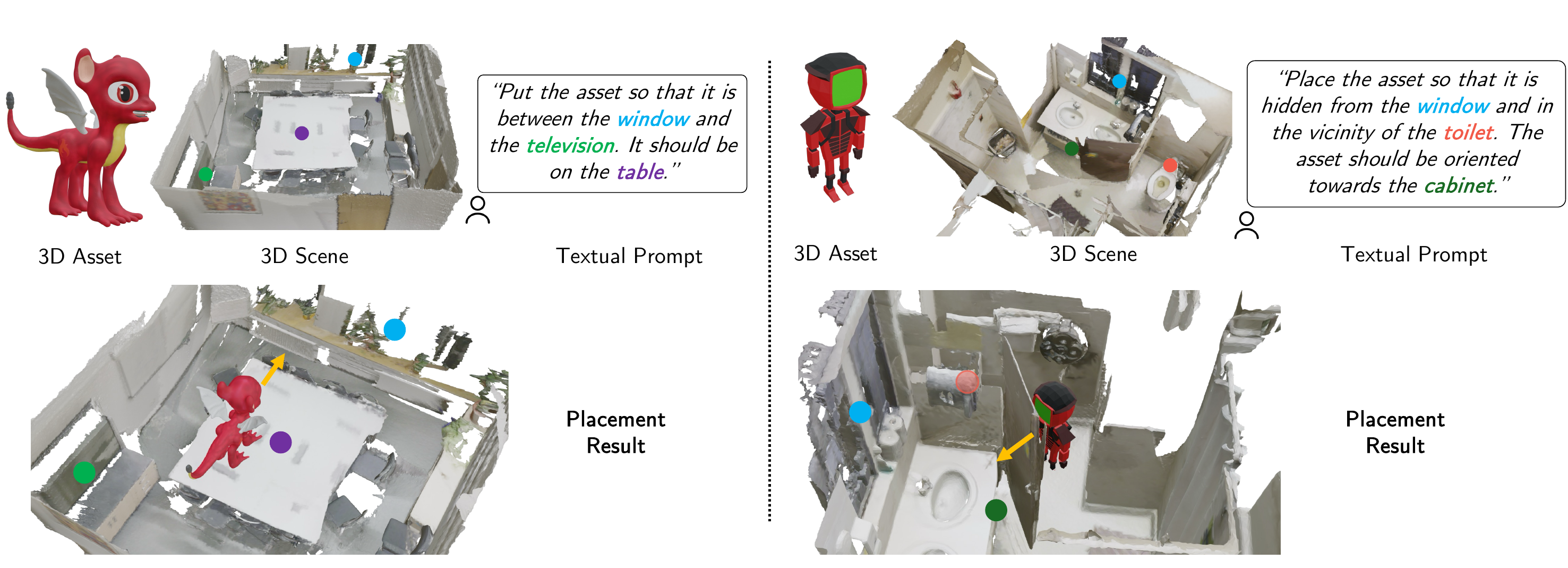}
\captionof{figure}{\textbf{Language-guided 3D Object Placement in Real 3D Scenes}: Given a text prompt, the task is to find a valid placement for an asset, requiring semantic and geometric understanding of the scene, the asset’s shape, and spatial relationships. Colored dots show referenced objects (for visualization only, not given to the model), and the yellow arrow indicates the predicted frontal direction.
}
\label{fig:teaser}
\end{center}
}]

\vspace{-1em}

\maketitle

\begin{abstract}
We introduce the task of Language‑Guided Object Placement in Real 3D Scenes. Given a 3D reconstructed point‑cloud scene, a 3D asset, and a natural‑language instruction, the goal is to place the asset so that the instruction is satisfied. The task demands tackling four intertwined challenges: (a) one‑to‑many ambiguity in valid placements; (b) precise geometric and physical reasoning; (c) joint understanding across the scene, the asset, and language; and (d) robustness to noisy point clouds with no privileged metadata at test time. The first three challenges mirror the complexities of synthetic scene generation, while the metadata‑free, noisy‑scan scenario is inherited from language-guided 3D visual grounding. We inaugurate this task by introducing a benchmark and evaluation protocol, releasing a dataset for training multi-modal large language models (MLLMs), and establishing a first nontrivial baseline. We believe this challenging setup and benchmark will provide a foundation for evaluating and advancing MLLMs in 3D understanding.
\end{abstract}

\section{Introduction}
\blfootnote{$^*$Work done during an internship at Niantic.}
\label{sec:intro}
At two to three years old, neurotypical children learn to follow two-step instructions like \textit{``Get your shoes and put them on the shelf''}~\cite{siegler2020children}. These tasks may appear simple, yet children need time to grasp basic vocabulary and to learn the physical affordances of both 3D objects and scene layout. Perhaps AIs could obtain similar capabilities. 

In this paper, we focus on the novel task of \emph{language-guided 3D object placement in a reconstructed real 3D scene}. As in the shoe example, the goal is to find a valid placement of the object among multiple configurations that satisfy the instruction. As shown in Figure~\ref{fig:teaser}, the placement must also respect the physical constraints of the space and of the 3D asset. Excelling at this task would unlock applications such as instructing a robot, through language, to move a real object to a new location. It is also relevant to augmented reality (AR). For instance, a shopper wearing an AR headset could use natural‑language commands to anchor a virtual product on a real‑world surface or reposition digital décor anywhere in the room.

MLLMs have recently shown strong performance on language-guided 3D scene understanding tasks, including visual question answering~\cite{hong20233d_llm, chen2024ll3da, Huang2023ChatSceneB3}, visual grounding~\cite{huang2024reason3d,zhu2024scanreason}, and synthetic scene generation~\cite{feng2023layoutgpt, yang2024holodeck, sun2025layoutvlm}. We study language-guided 3D asset placement in reconstructed scenes, a problem closest to grounding and to synthetic scene generation, yet distinct in that it requires addressing \emph{all} of the following challenges simultaneously:
\begin{enumerate}[label=\roman*.,  leftmargin=1.0em, labelsep=0.5em]
    \item \textit{Ambiguity of solutions}. 3D visual grounding typically admits a single correct match, whereas 3D placement is inherently one-to-many: multiple placements can satisfy the instruction~\cite{fisher2012example, chang2014learning}. This complicates both benchmarking and data construction, which must accommodate multiple valid answers.
    \item \textit{Intrinsic 3D reasoning}. Many constraints are geometric and cannot be resolved from 2D projections alone. For example, \textit{``place the asset in between the chair and the table, hidden from the window''} requires reasoning about free space and spatial relationships in 3D.
    \item \textit{No privileged information at test time}. In contrast with synthetic scene generation, we require just the reconstructed 3D scene. We are not given layouts~\cite{paschalidou2021atiss, sun2025layoutvlm}, scene graphs~\cite{zhou2019scenegraphnet}, object properties, or clean geometry~\cite{yang2024holodeck, huang2025fireplace} to aid with making a prediction.
    \item \textit{Joint reasoning over scene, asset, and language}. The asset’s size and shape restrict feasible placements; given the same scene and instruction, a large object has fewer valid locations than a small one. Among the valid options, the model must follow the user’s stated intent rather than default to common sense priors~\cite{yu2011make,Merrell2011interactive}.

\end{enumerate}
\vspace{0.2cm}
\noindent The highlight of this paper is the introduction of a challenging novel task, which we call \dataset{}: language-guided 3D placement on real scenes. To the best of our knowledge, no existing benchmarks, datasets, or methods directly address this problem. To advance research in this area, we make three key contributions, summarized here:
\begin{itemize}
    \item We introduce \dataset{}-benchmark for language-guided placement with 3,500 evaluation examples, each consisting of a real ScanNet scene~\cite{dai2017scannet}, a PartObjaverse-Tiny asset~\cite{yang2024sampart3d}, and a guiding prompt. Our evaluation protocol accounts for placement ambiguity, enabling fair comparison across methods.
    \item We present \dataset{}-dataset, a large-scale training dataset with 100,505 training and 2,566 validation examples, each annotated with all valid placements that can be used for training MLLMs. Like the benchmark, it uses ScanNet scenes and PartObjaverse-Tiny assets.
    \item We propose \network{}, a proto-method for this task built on recent 3D LLMs~\cite{huang2024reason3d}. It uses a modified form of spatial aggregation, an asset encoder, and rotation prediction, outperforming existing baselines.
\end{itemize}
\newpage

\begin{figure*}[ht!]
    \centering
    \includegraphics[width=1\textwidth]{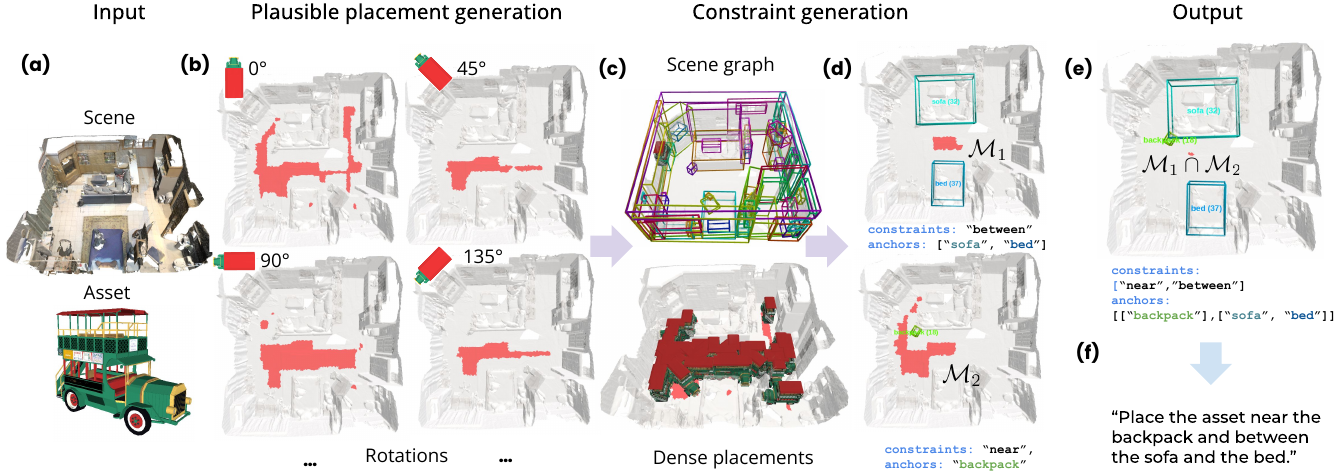}
    \caption{\textbf{\dataset{}-dataset creation}. Given a scene and an asset as input \textbf{(a) }the goal is to create a prompt \textbf{(f)} and corresponding mask $\mask$ of valid placements \textbf{(e)}. We start by finding the set of points which are physically plausible placements, shown in red in \textbf{(b)}. 
    We consider eight equally spaced rotation angles, which condition the valid placements. For this example, angle $0^{\circ}$ has more valid placements than $45^{\circ}$. To generate the language constraints, we use the ground truth scene graph \textbf{(c)}. Object anchors are selected from the scene graph and combined with relationship types to create a constraint and corresponding validity mask \textbf{(d)}. The different placement constraints are combined in the final output by intersecting the validity masks \textbf{(e)} given a mask of valid \textit{dense} placements.
    Based on each selected set of anchors and constraint relationships, a natural language prompt is created using templates (please, see supplemental for more details).
    }
    \label{fig:dataset_teaser_figure}
\end{figure*}

\section{Related Work}\label{sec:related}
\noindent\textbf{3D and Large Language Models.}
3D LLMs~\cite{ma2024llmsstep3dworld} are a subclass of MLLMs that jointly process point clouds/meshes and text. They typically align 3D features to the language space with lightweight projection layers or cross‑attention adapters, mirroring strategies from image-grounded MLLMs (VLMs)~\cite{liu2024visual,li2024llava_next_interleave,li2024llava_one_vision,chen2024spatialvlm,cai2024spatialbot,cheng2024spatialrgpt,lai2024lisa}. For grounding and QA, 3D-LLM~\cite{hong20233d_llm} fuses point clouds with 2D image features; Reason3D~\cite{huang2024reason3d} routes a pre-trained point encoder through a Q-Former~\cite{li2023blip} and a decoder guided by both the 3D input and the LLM output.  ScanReason~\cite{zhu2024scanreason} interleaves grounding and reasoning during inference. However, none of these systems addresses language-guided object placement, which requires reasoning about free space, spatial relations, and asset properties. We introduce a 3D LLM baseline designed to model these factors.

\noindent\textbf{3D object placement and scene generation.}
Early 3D layout work posed furniture arrangement as optimization over ergonomic or aesthetic constraints, sampling layouts that satisfy manually encoded rules and user constraints~\cite{yu2011make,Merrell2011interactive}. Data driven methods replaced hand tuning with spatial priors learned from example scenes, yielding \textit{placement masks} that reflect one-to-many valid locations~\cite{fisher2012example,chang2014learning,chang2017sceneseer}. Neural methods extend these priors to full indoor synthesis with scene graphs, transformers, or diffusion models~\cite{wang2019planit,zhou2019scenegraphnet,paschalidou2021atiss,feng2023layoutgpt,wei2023lego,yang2024scenecraft,tang2024diffuscene}, and hybrid systems mix learned priors with explicit geometric constraints~\cite{feng2023layoutgpt,yang2024holodeck}.  \cite{coyne2001wordseye} pioneered the use of language as a control signal for 3D scene generation. Later systems parse text into spatial constraints or scene graphs to infer feasible regions~\cite{chang2014learning,chang2015text,chang2017sceneseer,ma2018language}. Recent LLM and VLM based methods broaden this paradigm \cite{yang2024holodeck, tang2024diffuscene, fu2024anyhome, sun2025layoutvlm}. When it comes to real-world (\ie non-synthetic) scenes, most existing methods are image-based, predicting plausible 2D placement regions~\cite{liu2021opa,sharma2024octo,ramrakhya2024seeing,zhao2018compositing,zhu2023topnet}. RoboPoint~\cite{yuan2024robopoint} extends this by performing language-guided placement in images and then lifting the results to 3D using depth. Concurrent work FirePlace~\cite{huang2025fireplace} focuses on synthetic, clean environments. In contrast, we tackle language-guided placement in reconstructed real-world scenes, which requires intrinsic 3D reasoning, no privileged test-time annotations, and joint understanding of scene geometry, object shape, and instruction intent. Inspired by optimization methods, our dataset and benchmark ensure geometric feasibility for all candidate placements.

\noindent\textbf{Datasets for language-guided 3D tasks. }Most language and 3D datasets based on ScanNet~\cite{dai2017scannet} focus on tasks like text grounding~\cite{achlioptas2020referit3d,chen2020scanrefer,abdelreheem2024scanents3d,jia2024sceneverse}, VQA~\cite{azuma2022scanqa}, captioning~\cite{chen2021scan2cap}, and instruction following~\cite{zhang2024vla3D}. Larger real-scene corpora~\cite{embodiedscan,arnaud2025locate3drealworldobject,zhu20233dvistapretrainedtransformer3d} and synthetic embodied benchmarks~\cite{shridhar2020alfred,padmakumar2022teach,yang2024_3D_GRAND} expand this scope. Yet none address language-guided object placement with its one-to-many valid solutions and explicit valid placement enumeration; our ScanNet-based benchmark and dataset aim to fill this gap.

\vspace{-0.3cm}\section{Language-Guided 3D Object Placement}
\label{sec:taskAnddataset}

We introduce the task of language-guided 3D object placement on 3D reconstructed scenes.
Given the point cloud of scene, a 3D   asset, and text describing where the asset should be placed in the scene, the goal is to find a valid position and orientation for the asset that is physically plausible and adheres to the language prompt.

This task is inherently ambiguous because, in general, multiple valid placements exist. The multiple placements in Fig.~\ref{fig:dataset_teaser_figure} (c) demonstrate this ambiguity and illustrate the complexity of our task when compared with related tasks like object grounding, which typically has a single solution.
\\

\noindent\textbf{Simplifying assumptions.} Given the ambiguity and complexity of our task, we make some simplifying assumptions to make the problem tractable. First, we assume the vertical orientation of the scene is fixed and given by the Z-axis. We also assume we know the vertical orientation of the asset as well as its frontal direction.
The asset is always placed on a horizontal surface, and only the yaw angle is considered, \ie rotation around the vertical axis.

\subsection{Physical plausibility and language constraints}
Valid placement in this task must satisfy a core set of common constraints. The first, \textbf{physical plausibility}, requires that the object does not intersect the scene mesh and rests on a surface. This constraint is language-independent and always enforced.

Beyond physical plausibility, the language instruction specifies how the object should be placed in the scene. Placement is often relative to an “anchor”, a named object instance that must be inferred at test time.
In practice, ``language constraints” capture both semantic and physical aspects of the placement. These language constraints are organized into three distinct groups:

\noindent\textbf{Spatial constraints:} These constraints specify the object’s location relative to one or more scene anchors. This group includes:
 (i) \textit{near} and \textit{adjacent}: the object is positioned within a specified distance from an anchor.
 (ii) \textit{on}: the object should directly rest on top of an anchor.
 (iii) \textit{between}: the object must be placed between two anchors.
 (iv) \textit{above} and \textit{below}: the object is located above or below an anchor.

\noindent\textbf{Rotational constraint:} This constraint focuses on the orientation of the object relative to scene anchors. The object is positioned so that it faces the anchor.

\noindent\textbf{Visibility constraints:} The object is either within the anchor's line of sight (\textit{visible}) or hidden from it (\textit{not visible}).

A candidate placement is considered a valid placement if and only if it simultaneously satisfies every constraint in that prompt.

\noindent\textbf{Prompt creation}:We generate language prompts using a template-based system, where each constraint is expressed through predefined sentence templates. A random subset of constraints and anchor objects from ScanNet annotations is sampled for each prompt. Prompts that cannot be satisfied due to conflicting or overly restrictive constraints, such as ``place the asset on the table and below the desk", are discarded during a verification step. Template details are provided in the supplemental material.

\begin{figure*}[!htb]
    \centering
    \includegraphics[width=\linewidth]{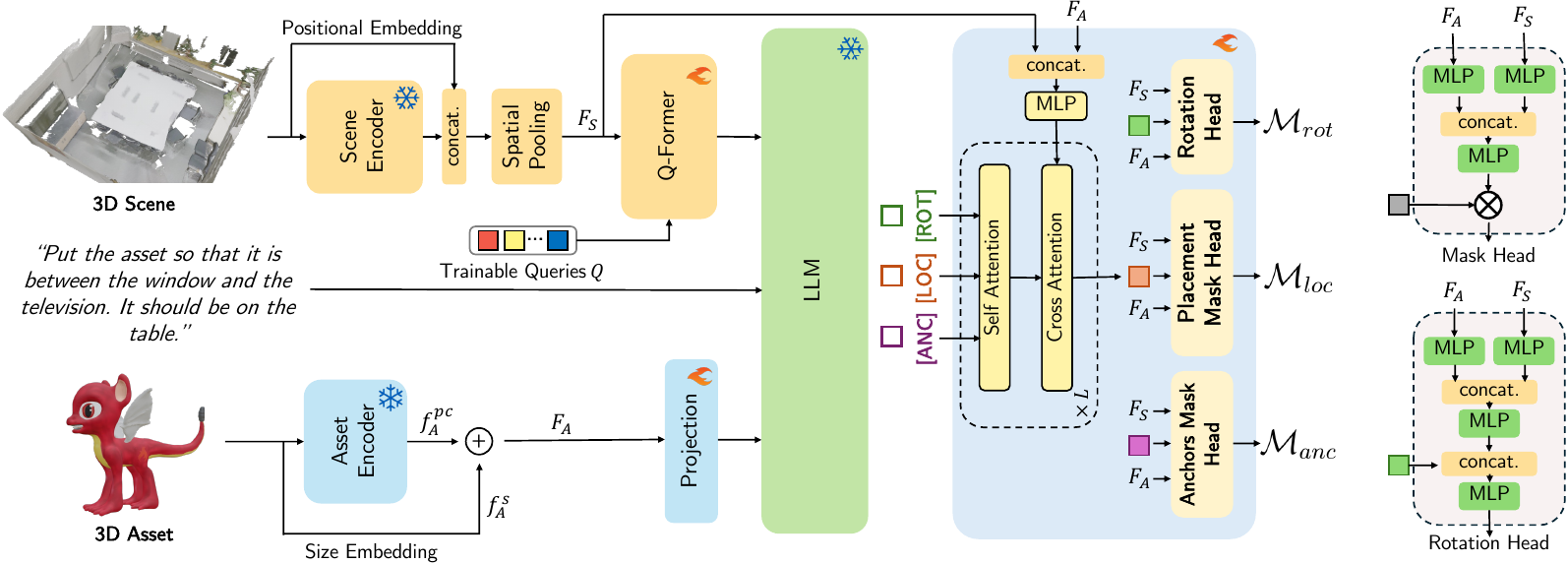}
    \caption{\textbf{\network{} overview.} A point encoder extracts features from the 3D scene, which are then complemented with positional embeddings. Spatial pooling reduces feature dimensions, and a Q-Former merges the pooled features with trainable queries $Q$ (Section \ref{sec:scene_encoding}). The asset is encoded into a single vector by using a pretrained asset encoder followed by max-pooling  (Section \ref{sec:asset_encoding}). This vector together with a size embedding is passed to a projection layer that aligns the features with the LLM space.
    The LLM take as input (i) the output of the Q-Former, (ii) the text prompt, and (iii) the projected asset features and predicts three special tokens $\anc$, $\loc$ and $\rot$. A transformer based decoder takes as input the features associated with the three special tokens and the pooled scene features and performs a few self and cross attention operations (Section \ref{sec:decoder}). Three heads produce the final outputs: $\mask_{loc}$ the valid placement mask; $\mask_{anc}$ an auxiliary mask that localizes the object anchors; and $\mask_{rot}$ a mask indicating which rotation angles are valid at each location.}
    \label{fig:method}
\end{figure*}
\vspace{-0.5cm}

\subsection{\dataset{}-benchmark}\label{sec:benchmark}

Each benchmark example consists of a 3D scene mesh, a 3D asset, and a language prompt comprising one or more 3D placement constraints. A placement method takes this triplet as input and predicts a placement, defined by a 3D translation vector $\mathbf{t}$ and a yaw angle $\alpha$. Our evaluation protocol checks whether each constraint and all constraints collectively are satisfied.
\subsubsection{Checking validity of each 3D constraint}
We use a rule-based system to verify if a predicted placement satisfies 3D constraints, using ground truth oriented bounding boxes from ScanNet for anchor objects. Each constraint type is evaluated based on its specific properties, often allowing small deviations via thresholds.

\noindent\textbf{Physical plausibility:} We use \texttt{trimesh}~\cite{trimesh} to check for intersections between the object and the scene and whether the object is placed on a surface. 

\noindent\textbf{Spatial constraints:} For the \textit{near} and \textit{adjacent} constraints, we compute the distance from the posed asset to the anchor object. We check the \textit{on}, \textit{above}, and \textit{below} relationships by comparing the value of the z-coordinate of the placed object with the z-coordinate of the anchor object. For the \textit{between} relationship, we check if the placed asset is close to a line connecting the centers of the two anchor objects.

\noindent\textbf{Rotational constraint:} We compute a cone around the frontal vector of the posed asset and check that the anchor object intersects with that cone.

\noindent\textbf{Visibility constraint:} To determine visibility from a given anchor, we render the object and scene from a camera at the anchor point facing the object, and check if any pixels in the image correspond to the object. %

More details are available in the supplemental material.

\subsubsection{Benchmark metrics}
\label{benchmark_metrics}
To evaluate placement performance, we compute metrics that capture constraint validity overall and by subgroup:
\begin{itemize}
    \item \noindent\textbf{Global Constraint Accuracy:}  The percentage of all constraints (across all groups) that are correctly satisfied over the entire dataset. It provides a holistic measure of the overall placement quality.
    \item \noindent\textbf{Complete Placement Success:} The percentage of perfect valid placements, where every constraint—including physical plausibility—is satisfied. This is a strict metric that reflects the robustness of the placement method under full constraint satisfaction.
    \item \noindent\textbf{Language Adherence Success:} The percentage of placements that satisfy all language-based 3D constraints, excluding physical plausibility. It measures whether the model fully adheres to the language instructions.
    \item \noindent\textbf{Subgroup Metrics:} In addition to the overall metrics, we report accuracies across constraint groups.
\end{itemize}

\subsubsection{Benchmark statistics}
The benchmark contains 3,500 evaluation examples, combining a total of 142 different scenes from ScanNet~\cite{dai2017scannet} and 20 different assets from the PartObjaverse-Tiny dataset~\cite{yang2024sampart3d}. Statistics for the number of constraints and type of constraints are shown in Table~\ref{tab:benchmark_statistics}.

\begin{table}[h!]
\small %
\centering
\begin{tabular}{@{}l r@{\hskip 20pt} |lr@{}}
\toprule
{Constraints per sample} & {\#} & {Type} & {\#} \\
\midrule
One & 900 & Spatial & 4,208 \\
Two & 1,871 & Rotational & 1,503 \\
Three or more & 729 & Visibility & 1,210 \\
\bottomrule
\end{tabular}
\caption{Benchmark statistics for the number and types of language constraints per sample. Physical plausibility is evaluated in all samples and thus excluded from this table.}
\label{tab:benchmark_statistics}
\end{table}

\subsection{\dataset{}-dataset: Training dataset}\label{sec:dataset}

Although our benchmark protocol allows offline method evaluation, we need a practical, less computationally costly approach to create a large-scale dataset for training, especially for obtaining the full set of valid placements.

Here we describe \dataset{}-dataset, our training dataset for the task of guided placement.
The dataset consists of 100,505 training examples, sourced from 565 distinct ScanNet scenes and 20 unique assets. It includes a total of 83,530 spatial constraints, 34,445 rotational constraints, and 18,746 visibility constraints. Among these examples, 65,586 contain a single constraint, 26,395 have two constraints, and 8,524 include three or four constraints. Used throughout this paper, this training dataset is a subset, for practical purposes, of the \dataset{}-dataset-full corpus we are also sharing. \dataset{}-dataset-full has $\sim$$4$M examples: the 565 scenes x 140 objects x 50 prompts.%

\noindent\textbf{Dataset parametrization} We denote the point cloud of the scene as $\PC \in \mathbb{R}^{N \times 6}$, where each point $\point_i, i \in \{0, ..., N-1\}$ contains the 3D position for that point, as well as color information.
Given a scene, an asset, and a prompt, we represent the set of valid ground truth placements for the asset as a binary mask $\mask$ defined over the point cloud of the scene, associating a label $m_i \in \B$ to each 3D point $\point_i$. 
For each point $i$ with label $m_i = 1$, \ie a valid placement, we also define a binary mask over a discretized set of yaw angles indicating if the angle is valid for that specific location: $\mathbf{\alpha}_{i} = \{\alpha^{y}_{i} \in \B | y= 0, ..., 7\}$, where each $y$ corresponds to a $45^\circ$ interval.
Note that there is a fixed transform between the parametrizations used in the benchmark and the training dataset. While for the benchmark we parametrize the position of the center of the asset, for the training set, we consider contact points between the scene geometry and the asset's bottom surface. 

\noindent\textbf{Computing valid placement masks}
We create the valid placement masks $\mask$ by using a combination of the rule-based system defined above and a few approximations to make it more efficient. More details on the approximations are available in the supplemental material. We treat each constraint independently, obtaining a valid mask per constraint $\mask_{c}$ with $c \in \mathcal{C}$, where $\mathcal{C} = \{\text{physical}, \text{spatial}, \text{rotational}, \text{visibility}\}$. The final mask is given by the intersection of all the constraints that apply to that example, so
\begin{equation}
\label{eq:valid_placement}
\mask \;=\; \bigcap_{c \in \mathcal{C}} \mask_c.
\end{equation} %
For the physical plausibility constraint we use a set of heightmaps to capture the different horizontal surfaces of the scene. We then compute the asset height and footprint and, for each point on a horizontal surface, check if the placement is valid.
For the visibility constraint we use the same procedure as the benchmark, but use two approximations for efficiency: the asset is replaced by its bounding box, and a fixed rotation angle is used.

\vspace{0.3cm}
\section{\network{}: Method Description}
\label{sec:method}

\textbf{Background.} We briefly introduce Reason3D \cite{huang2024reason3d} as our method builds upon it. Given a textual prompt and a colored point cloud $\PC \in \R^{N \times 6}$ as input, Reason3D performs dense 3D grounding, finding all the points in the point cloud that satisfy the prompt. A point encoder~\cite{sun2023superpoint} extracts features $F_X \in \R^{N \times d}$ from the input point cloud, where $d$ is the feature dimension.  These are aggregated into superpoints~\cite{landrieu2018large} obtaining superpoint features $F_S \in \R^{M \times d}$, with $M \ll N$, reducing the overall complexity. %

Next, the superpoint features $F_s$ are projected into the embedding space of an LLM via a Q-Former block \cite{li2023blip}. This model updates the learnable query vectors $Q$, resulting in $Q'$. From $Q'$ and the input text, the LLM generates a response containing two special tokens, namely $\loc$ and $\seg$. These tokens guide the model in two stages: coarse localization followed by precise mask prediction.

In practice, the Reason3D method uses a single token, $\loc$, for datasets that contain small scenes, such as ScanNet, since hierarchical subdivision is not required. We will describe their method using this simplified version.

Finally, the last-layer embeddings associated to $\loc$ are first projected via an MLP and then given as input to the Mask Decoder, which performs cross-attention \cite{vaswani2017attention} with $F_s$. The decoder produces an object-level binary segmentation mask over superpoints, which is upsampled into $\mask_{loc} \in \B^{N}$ to provide a segmentation mask on the full point cloud.

Figure \ref{fig:method} provides an overview of our method. In the following subsections, we detail our approach and emphasize the key modifications to the Reason3D architecture necessary for addressing \textit{language-guided placement instead} of standard 3D visual grounding.

\subsection{Scene encoding}\label{sec:scene_encoding}
Similarly to Reason3D, we use the point encoder from \cite{sun2023superpoint} to extract features $F_X \in \R^{N \times d}$ from the 3D scene.
We use an additional positional embedding feature $F_X^{pos} \in \R^{N \times d^{*}}$, for points in the point cloud, encoding their location, which is concatenated with the previous features. 

\noindent\textbf{Spatial pooling. }
Reason3D uses Superpoints~\cite{landrieu2018large} to reduce computational complexity and memory usage by pooling individual point features into a single feature per superpoint. Although effective for their task, this coarse representation limits performance for our placement task.

 For example, Superpoints will generally cluster all points belonging to horizontal or vertical surfaces --such as floors, tabletops and walls-- into single Superpoints, which is clearly undesirable for accurate 3D placement of assets. To address this, we instead use uniform spatial pooling to aggregate features.
Specifically, we use farthest point sampling~\cite{han2023quickfps} to select $M$ center points, then assign each point in the cloud to its nearest center based on Euclidean distance. This approach keeps the method computationally efficient while maintaining sufficient granularity for accurate asset placement. The level of granularity reflects a trade-off between computational cost and the model's ability to reason over fine-grained geometric details. This trade-off is evident in Table~\ref{tab:ablations_table}, where comparing row~\ref{row:reason3D} with row~\ref{row:uniform_aggregation} shows that finer sampling enables better spatial reasoning. Please see supplementary material for a visualization.

Our spatially aggregated features $F_S$ are passed as input to the Q-Former block \cite{li2023blip}, which also takes as input a set of trainable queries and learns to project the features into the LLM embedding space.

\subsection{Asset encoding}\label{sec:asset_encoding}
When compared with other tasks, our language-guided placement task has an additional input, the 3D asset point cloud. We encode the asset using a Point-BERT encoder \cite{yu2021pointbert} trained on the Objaverse \cite{Deitke2022ObjaverseAU} dataset. This encoder predicts a sequence of feature vectors that are max-pooled to obtain a single feature embedding.

Encoding the scale of the input asset is essential to facilitate a valid placement. 
Since the asset encoder assumes a normalized point cloud in a unit sphere, we separately encode the size of the asset by taking the asset's dimensions in the X, Y, and Z axes.
The $F_{A}$ feature for the asset is a combination of the asset encoding and scale embeddings and is projected to the LLM space using an MLP.

\subsection{Placement decoder}\label{sec:decoder}

We instruct our LLM to output three special tokens, namely a $\loc$ token, an $\anc$ token, and a $\rot$ token. 
The features associated with the three special tokens are passed as input to the decoder, where they undergo a few self-attention layers.
These are followed by a few cross-attention layers between the updated token features and the asset features $F_A$ and the pooled scene features $F_S$.

Each individual head takes the feature of the associated token after attention, the asset feature $F_A$, and the scene feature $F_S$ and predicts the corresponding output.
The \textbf{Placement Mask Head} takes the $\loc$ token embedding and predicts $\mask_{loc} 
\in \Bint^N$, a mask over the scene point cloud encoding the regions where the input asset can be placed satisfying the input prompt. The \textbf{Rotation Head} takes the $\rot$ token embedding and predicts $\mask_{rot}\in \Bint^{N \times 8}$ indicating for each point in the point cloud, the validity of a discretized set of rotation angles. 
Finally, the \textbf{Anchors Mask Head} takes the $\anc$ token and predicts $\mask_{anc} \in \Bint^N$, a mask encompassing the masks of all the anchor objects. This is used only as an auxiliary task, to help the network identifying anchors in the prompt. The head architectures are depicted on the right side of Figure~\ref{fig:method}.

\subsection{Losses}
We use a combination of Binary Cross Entropy (BCE) and Dice \cite{Sudre2017GeneralisedDO} losses when comparing a ground truth mask $\bar{\mask}$ with a predicted mask ${\mask}$, so
\begin{equation}
\Lo_{seg}(\bar{\mask}, \mask) = \text{BCE}(\bar{\mask}, \mask)  + \text{Dice}(\bar{\mask}, \mask).
\end{equation}

\noindent The loss for the rotation prediction is given by
\begin{equation}
\mathcal{L}_{rot} = \text{BCE}(\bar{\mask}_{rot}, \mask_{rot}),
\end{equation}
where $\bar{\mask}_{rot} \in \mask_{rot}\in \B^{N \times 8}$ is the ground truth indicator mask for valid rotation angles, per point in the scene point cloud.

The loss for the LLM is a cross-entropy loss, comparing the ground truth text $\bar{Y}$ with the predicted text $Y$: 
$\Lo_{L} = \text{CE}(\bar{Y}, Y)$. Note that the ground truth text $\bar{Y}$ for our task, follows a simple format, \eg ``Sure, it is [LOC][ANC][ROT]'', since %
the LLM is not required to predict articulated responses or explain placement decisions. Instead, the information useful for placement should be encoded in the embeddings for the special tokens. Finally, our total loss is defined as
\begin{equation}
 \Lo =
 \Lo_{seg}(\bar{\mask}_{loc}, \mask_{loc}) + \Lo_{rot} + \Lo_{seg}(\bar{\mask}_{anc}, \mask_{anc})
+ \Lo_{L}.
\end{equation}
\begin{table*}[htbp]
    \centering
    \normalsize
    \setlength{\tabcolsep}{3pt} %
    \renewcommand{\arraystretch}{0.95} 
    \resizebox{\textwidth}{!}{%
    \begin{tabular}{lcccccc cccc c ccc}
        \multicolumn{7}{c}{Method ablation} & \multicolumn{4}{c}{Subgroup metrics} &
        \multicolumn{3}{c}{Global metrics}& \\
        \cmidrule(lr){1-7} \cmidrule(lr){8-11} \cmidrule(lr){12-14} 
        Name & \makecell{Spatial\\aggr.} & \makecell{Pos.\\emb.} & \makecell{Asset\\encoder} & \makecell{Anchor\\pred.} & \makecell{Rot\\pred.} & \makecell{Decode\\asset} 
             & Physical & Spatial & Rotational & Visibility 
             & \makecell{Language\\adherence\\success} 
             & \makecell{Global\\constraint\\accuracy} & \makecell{Complete\\placement\\success} \\
        \midrule
        \multicolumn{7}{l}{Baseline --- OpenMask3D \cite{takmaz2023openmask3d} + rules} & \textbf{61.6} & 28.6 & 6.5 & 53.4 & 21.8 & 29.2 & 11.7\\
        \multicolumn{7}{l}{Baseline --- OpenMask3D~\cite{takmaz2023openmask3d} +  LLM }  & 5.8 & 35.3 & 10.5 & \textbf{61.5} & 18.4 & 26.7 & 1.6 \\
        \row{row:reason3D} --- Reason3D~\cite{huang2024reason3d}       & Superpoints & --        & text     & --         & --        & --  & 53.9 & 37.5  & 6.6  & 57.0 & 18.1 & 44.8 & 13.2 \\
        \midrule
        \row{row:uniform_aggregation}  & uniform     & --        & text     & --         & --        & --  & 56.3 & 47.4	& 8.6	& 56.1 & 18.4 & 48.9 & 10.1 \\
        \row{row:pos_embedding}   & uniform     & \checkmark & text    & --         & --        & -- & 58.6 & 49.7	& 7.9	& 59.1 & 20.0 & 50.4 & 10.9 \\
        \row{row:pointbert}     & uniform     & \checkmark & PointBert& --         & --        & -- & 57.9 & 47.1	& 7.5	& 58.5 & 17.2 & 49.3	& 9.6 \\
        \row{row:anchors_head}  & uniform     & \checkmark & PointBert& \checkmark & --        & -- & 57.5 & 54.3	& 10.0	& 58.8 & 22.2 & 42.5	& 12.3 \\  
        \row{row:rotation_head}& uniform     & \checkmark & PointBert& \checkmark & \checkmark& -- & 55.6 & 50.8 & 14.7 & 57.8 & 20.8 & 51.0 & 11.4 \\       
        \row{row:ours_final} --- \small\network{} & uniform     & \checkmark & PointBert& \checkmark & \checkmark& \checkmark & 58.8 & \textbf{56.6} & \textbf{17.3} & 61.2 & \textbf{25.9} & \textbf{54.9} & \textbf{15.0} \\
        \bottomrule
    \end{tabular}%
    }
    \caption{\textbf{Quantitative results:} We compare our full method with variations where some components are removed. The results validate our design choices, and they show improvements over OpenMask3D~\cite{takmaz2023openmask3d} with rule-based and LLM-based asset placement and Reason3D~\cite{huang2024reason3d}.}
    \label{tab:ablations_table}
\end{table*}

\vspace{-0.7cm}
\subsection{Inference}
At inference time, our method takes the network predictions for placement, $\mask_{loc}$, and rotation, $\mask_{rot}$, and extracts a single valid placement by finding the point in the point cloud with the maximum value in $\mask_{loc}$:
$\hat{\point} = \text{argmax}_{m \in \mask_{loc}}m$.
We apply a fixed offset to point $\hat{\point}$, half the asset height, to get the predicted 3D translation vector $\hat{t}$. This is due to the differences in parametrization between the training dataset and the benchmark. To predict the rotation angle, we use $\mask_{rot}^{\hat{\point}} \in \Bint^8$, which encodes the validity of discretized rotations for $\hat{\point}$. The predicted angle $\hat{\alpha}$ is obtained by taking the $\mathrm{argmax}$ over this vector.

\newpage\begin{figure*}[!htb]
    \centering
    \vspace{-0.5cm}
    \includegraphics[width=0.98\linewidth]{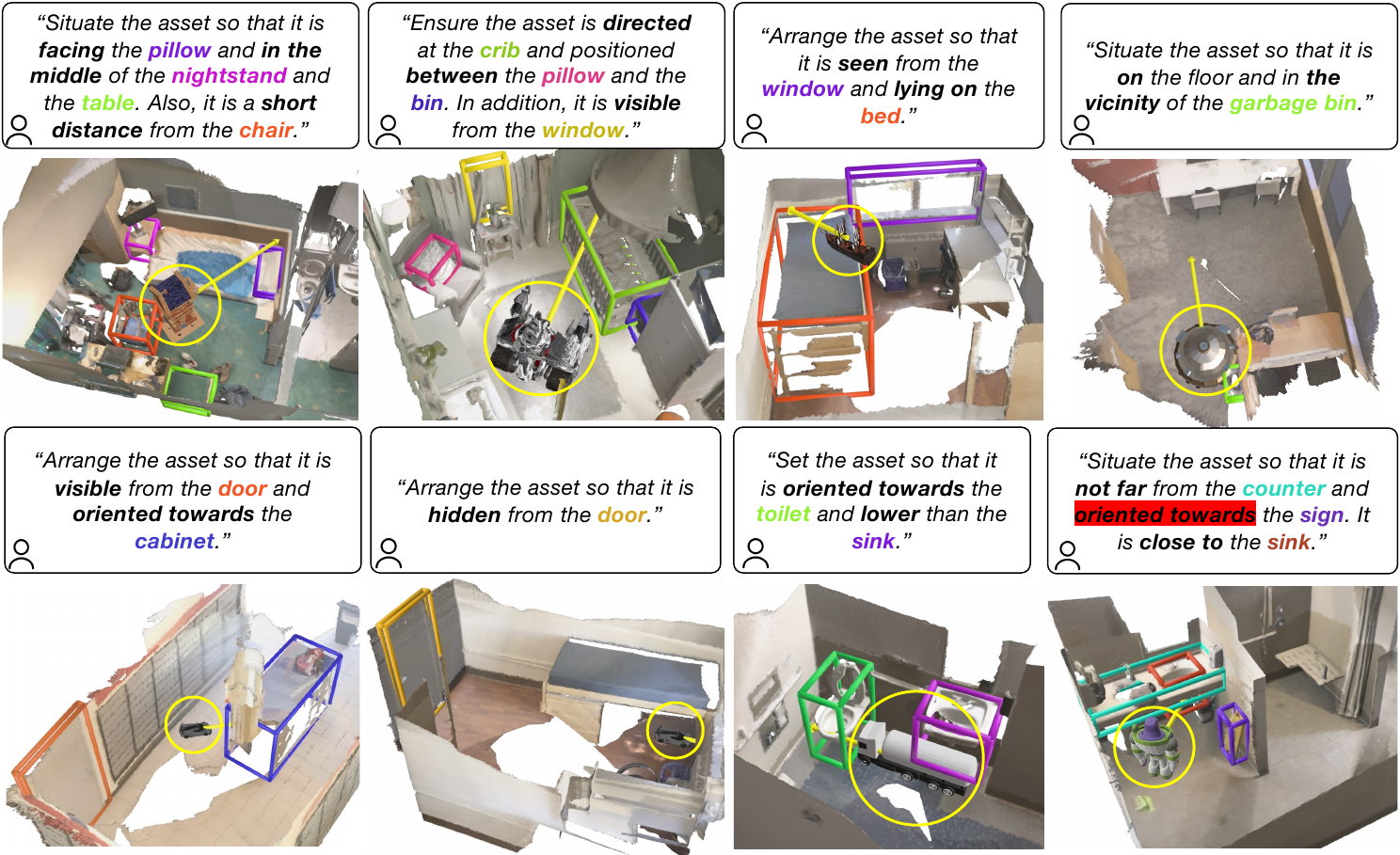}\vspace{-0.2cm}
    \caption{\textbf{Qualitative benchmark results}. Colored highlights indicate anchors referenced in the textual prompts (predictions are generated entirely from point clouds, with anchor information provided only as text). The asset position is marked with a yellow circle, and a yellow arrow denotes the frontal orientation. Our method successfully follows language instructions and meets the specified constraints. The top-right example illustrates a placement that satisfies constraints but slightly intersects with the scene mesh. The bottom-right example demonstrates a failure case where one constraint is not met (highlighted in red).}
    \label{fig:qualitative}\vspace{-0.3cm}
\end{figure*}

\section{Experiments}
\label{sec:experiments}
We validate our method \network{} for the task of language-guided object placement on the benchmark described in Section~\ref{sec:benchmark}.  Our metrics, described in Section \ref{benchmark_metrics}, measure prediction validity. All values are percentages, where higher is better. Implementation details are in the supplemental material.

\subsection{Quantitative results}
In the absence of prior work on language-guided 3D object placement in real scenes, we implemented two baselines by integrating OpenMask3D \cite{takmaz2023openmask3d}, an open vocabulary grounding method, with two different placement strategies: (1) a rule-based optimization approach, similar to the one used during training data generation in Section~\ref{sec:dataset}, and (2) an LLM-based system that uses GPT-4o, with the estimated scene graph provided as part of the input prompt. We evaluate both baselines against our proposed method. Since OpenMask3D requires explicit object queries, we use ground truth anchor descriptions (rather than full placement instructions) to retrieve relevant regions. Due to its frequent failure to accurately detect floor regions, we substitute in ground truth floor masks, while other anchor objects are selected based on the highest similarity scores. Once anchor masks are obtained, asset placement is performed either by the rule-based strategy or inferred via the LLM. 

Table~\ref{tab:ablations_table} presents both baseline comparisons and ablation results. Our method, row~\ref{row:ours_final}, consistently outperforms both baselines across all overall evaluation metrics. The LLM-based system often produces physically implausible placements, primarily because it lacks direct access to the 3D geometry of the scene. In contrast, the rule-based system, which leverages both asset and scene meshes, can produce more plausible placements, albeit at the cost of expensive collision checks during inference. By comparison, our method's end-to-end design eliminates the need for such costly test time operations, making it more scalable for large and complex scenes. Additionally, we observe that our method more accurately follows the user’s language instructions than either baseline. Finally, the relatively low scores across all methods under the strictest evaluation metric, Complete Placement Success, which requires both physical plausibility and full adherence to all language constraints, highlight the inherent difficulty of the task.

\subsubsection{Ablations}
Table~\ref{tab:ablations_table} also shows results for different ablations of our method.
We start with an adaptation of the Reason3D~\cite{huang2024reason3d} model to our task. One by one, we incrementally modify it using our novel components. Each row in the table introduces a single new modification, as compared to the previous row. We evaluate and report the model’s performance until we reach \network{}, our final method. 
All models are trained on our training dataset \dataset{}-dataset. For the methods that do not predict rotation (rows \ref{row:reason3D}, \ref{row:uniform_aggregation}, \ref{row:pos_embedding}, \ref{row:pointbert}, and \ref{row:anchors_head}) we set the predicted rotation angle to $0$.
We describe the different variants below.

\noindent\textbf{\ref{row:reason3D}}. The asset dimensions are encoded in text and provided as part of the prompt: ``\textit{The asset dimensions are X Y Z cm}", where X, Y, and Z are integer values in cm.

\noindent\textbf{\ref{row:uniform_aggregation}}. This variant uses our proposed uniform spatial pooling approach instead of the original superpoints pooling.

\noindent\textbf{\ref{row:pos_embedding}}. Positional embedding features $F_X^{pos}$ for points in the point cloud are added to the scene encoding.

\noindent\textbf{\ref{row:pointbert}}. We incorporate the asset encoder instead of only providing the asset dimensions in the text prompt to the LLM. 

\noindent\textbf{\ref{row:anchors_head}}. We introduce the anchor prediction auxiliary loss. In \cite{abdelreheem2024scanents3d}, predicting anchor objects leads to better 3D visual grounding. We find that this holds for our task as well. 

\noindent\textbf{\ref{row:rotation_head}}. We introduce a rotation prediction head, enabling the model to predict also the rotation mask $\mask_{rot}$.

\noindent\textbf{\ref{row:ours_final}}. This variant defines our final method, where the asset feature $F_A$ is added as an additional input to the placement decoder. This integration helps the decoder reason more effectively about the asset's geometry in relation to the scene.

The results in Table \ref{tab:ablations_table} validate our design choices. Using spatial aggregation instead of superpoints improves over almost all metrics (compare row \ref{row:uniform_aggregation} with row \ref{row:reason3D}). The inclusion of the anchor prediction head as an auxiliary sub-task also improves performance  (row \ref{row:anchors_head} vs row \ref{row:pointbert}). Finally, the use of our rotation head combined with passing the asset encoding as input to the decoder gives our final best-performing method (row \ref{row:ours_final}, which we use in the qualitative results). 
\vspace{-0.1cm}
\subsection{Qualitative Results}
\vspace{-0.1cm}
In Figure~\ref{fig:qualitative}, we show the results of our method \network{} on benchmark examples, demonstrating its ability to follow language instructions and satisfy constraints. While most placements are accurate, some cases exhibit minor intersections with the scene mesh or constraint failures. %

\vspace{-0.1cm}
\section{Limitations and Future Work}\vspace{-0.1cm}

Our novel task formulation currently has several limitations. First, we focus exclusively on placing objects on horizontal surfaces. Extending this to support arbitrary contact points would enable broader applications, such as hanging a clock on a vertical wall. Second, our dataset and method do not address inconsistencies in language guidance, where instructions may not align with the actual scene. Additionally, both the dataset and the benchmark rely on synthetic rule-based optimization without human verification. This limits annotation quality, especially in edge cases, and consequently affects both prediction accuracy and evaluation reliability.
Despite these limitations, we believe our work lays the groundwork for further research in this area.  Our method can also be seen as a specialist model, as it is trained and evaluated solely on the guided placement task. Exploring how to integrate this task into a more generalist framework remains an important direction for future work.
\vspace{-0.2cm}
\section{Conclusion}\label{sec:conclusions}\vspace{-0.1cm}
We introduced a new task, benchmark, and dataset for language-guided object placement in real 3D reconstructed scenes, connecting natural language understanding with spatial reasoning over both scenes and assets. The benchmark is designed to reflect the inherent ambiguity of placement, where multiple valid solutions are possible. We also proposed a baseline method built on recent advances in 3D LLMs, supported by ablation studies that highlight the impact of key architectural and design choices. Benchmark results show considerable room for improvement, and we hope this work provides a foundation for advancing research in 3D spatial reasoning.

\section{Acknowledgments}

We thank Jakub Powierza and Stanimir Vichev for their help with the experimental infrastructure.
This work was partially supported by funding from King Abdullah University of Science and Technology (KAUST) – Center of Excellence for Generative AI, under award number 5940.

{
    \small
    \bibliographystyle{ieeenat_fullname}
    \bibliography{main}
}

\maketitlesupplementary

\section{Additional details on the training dataset}

\begin{figure*}[ht!]
    \centering
    \includegraphics[width=1\textwidth]{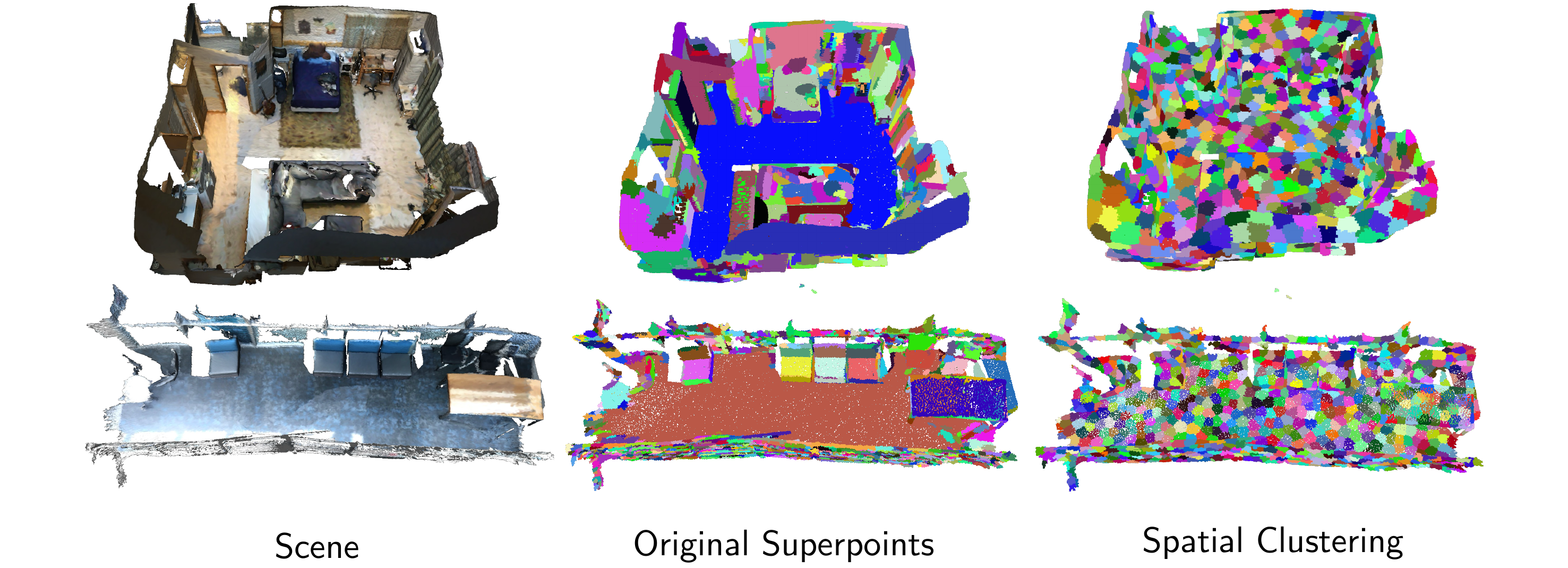}
    \caption{Comparison of our spatial pooling vs the superpoints used in \cite{huang2024reason3d}. Our regions are more local and more adequate to the task of object placement.}
    \label{fig:superpoint}
\end{figure*}

\subsection{Training dataset creation}

We give some details on the training set creation, particularly how the physically plausible constraint and visibility constraint are computed.

\paragraph{Spatial constraints}
 Each constraint uses geometric criteria on 3D oriented bounding boxes and is governed by the following parameters. We use the same values both in the training dataset and the evaluation benchmark:

\begin{itemize}

    \item \textbf{``near"}: asset in a proximity of the anchor object. The threshold distance is proportional to the size of the room (1\%).
    \item \textbf{``adjacent"}: asset close to the anchor object. We set a tolerance distance of 3 cm. 
    \item \textbf{``above" / ``below"}: asset vertically aligned above / below the anchor object. Vertical Intersection over Minimum (IoM) $>=$ 0.5 and a minimum of 1 cm above/below the anchor.
    \item \textbf{``on"}: resting on top of the anchor object, considering vertical stacking and size constraints: vertical IoM $>=$ 0.5 and a tolerance for vertical distance of 1 cm.
    \item \textbf{``between"}: Determines if the asset object lies between two anchor objects in both xy and z planes. Parameters: between IoM (0.5) in projection planes (xy and z). Overlap threshold (0.3): maximum IoM that ensures the asset does not overlap excessively with either object. Distance threshold: filters anchors beyond 1.5 m
\end{itemize}

\paragraph{Rotational constraint.} The ``facing" constraint determines which objects an asset is oriented towards in a 3D scene by evaluating directional alignment, proximity, and spatial overlap. It uses the asset's front direction to identify candidate objects within its field of view. We use a maximum distance threshold of 2 meters, an angular threshold of 30 degrees and an IoM for lateral overlap of 0.5.
\paragraph{Physically plausible constraint}

The first constraint that we consider is whether an object can physically be placed at a particular location in a scene.
To compute valid placements in a scene efficiently, we make use of a heightmap based representation, where we raycast the mesh from above using a grid of rays with a predefined resolution, and store all points of intersection.
Next, we create a set of heightmaps, where each cell represents a different ray, each layer represents a different intersection per ray, and the value is the height of the intersection point. 
We construct the first heightmap using the intersection points with the minimum height per ray. 
Each subsequent heightmap will contain either the next intersection point for each ray, or if there are no remaining points for a cell, will contain the maximum intersection point. 
Additionally, we raycast each asset from above to obtain an asset heightmap and 2D bounding box per possible rotation. 
Given these heightmaps, we check for physical plausibility by:
\begin{itemize}
    \item Extracting overlapping patches of the mesh heightmap, with patch size equal to the asset bounding box
    \item Extract minimum height and maximum height of the heightmap for each patch, using the asset heightmap to generate a mask. If this differs by more than 10cm, this point is not valid
    \item Check that the asset can also fit in the Z direction using the next heightmap - is the asset height for this cell less than the height of the next surface
\end{itemize}
If these 2 conditions are true, we deem a location to be physically plausible. 
Finally, we generate labels for the mesh vertices by assigning them the labels of their nearest location in the heightmap.

\paragraph{Visibility constraint}
Our visibility constraint determines whether an asset is visible from a specific location in the scene, which corresponds to one of the object anchors. To evaluate this, we use mesh rendering. To assess visibility efficiently, we first place the asset in a physically feasible position. Instead of rendering the full asset mesh, we approximate it using a simple cuboid with the same dimensions as the asset bounding box, reducing computational overhead, we also consider only 1 rotation for the asset. The virtual camera’s position is then determined by computing the centroid of the vertices associated with the anchor. The camera center is set to the vertex within the anchor that is closest to this centroid, and the camera is oriented to face the asset.

We then render the scene and check whether any pixels from the asset's bounding cuboid appear in the rendered image. This process is repeated for all valid asset placements across all scenes. The virtual camera locations correspond to TVs, doors, and windows. When multiple instances of the same object class exist in a scene, we select the largest instance.

We use a virtual camera with a field of view (FOV) of $60^\circ$ and we render images at a resolution of $64\times64$ pixels.

In the benchmark, we follow the same procedure as stated above for generating the training data with 2 differences: we render the original asset mesh instead of the cuboid and render images at a resolution of $256\times256$ pixels.

\subsection{Templates for prompts}
We report the templates used to generate placement instructions.

\begin{lstlisting}[language=YAML,  backgroundcolor=\color{backgroundcolor}
]
relationships:
  - name: plausible
    templates:
    - in a plausible location
    - in a sensible location
    - in a reasonable spot
    - in a suitable position
    - in a feasible area
    - somewhere stable within the scene
    - at a steady spot in the scene
    - in a secure location within the scene
    - in a firm position in the scene
    - in an area that suits the scene's layout
  - name: adjacent
    templates:
      - adjacent to the anchor_class
      - next to the anchor_class
      - beside the anchor_class
      - right beside the anchor_class
      - alongside the anchor_class
      - abutting the anchor_class
  - name: between
    templates:
      - between the anchor1_class and the anchor2_class
      - in the space between the anchor1_class and the anchor2_class
      - positioned between the anchor1_class and the anchor2_class
      - in the middle of the anchor1_class and the anchor2_class
  - name: facing
    templates:
      - facing the anchor_class
      - directed at the anchor_class
      - pointing towards the anchor_class
      - oriented towards the anchor_class
      - looking at the anchor_class
      - angled toward the anchor_class
      - turned towards the anchor_class
  - name: near
    templates:
      - near the anchor_class
      - close to the anchor_class
      - in the vicinity of the anchor_class
      - not far from the anchor_class
      - within reach of the anchor_class
      - a short distance from the anchor_class
  - name: on
    templates:
      - on the anchor_class
      - resting on the anchor_class
      - placed on the anchor_class
      - sitting on the anchor_class
      - lying on the anchor_class
  - name: above
    templates:
      - above the anchor_class
      - over the anchor_class
      - higher than the anchor_class
      - up above the anchor_class
  - name: below
    templates:
      - below the anchor_class
      - under the anchor_class
      - beneath the anchor_class
      - underneath the anchor_class
      - lower than the anchor_class
      - situated under the anchor_class
      - right below the anchor_class
  - name: is_visible
    templates:
      - visible from the anchor_class
      - in view of the anchor_class
      - within sight of the anchor_class
      - seen from the anchor_class
      - not obstructing the view to the anchor_class
      - keeping the view to the anchor_class clear
      - positioned to avoid blocking the anchor_class
      - allowing an unobstructed view of the anchor_class
  - name: not_visible
    templates:
      - not visible from the anchor_class
      - out of sight of the anchor_class
      - hidden from the anchor_class
      - obstructing the view to the anchor_class
      - blocking the view to the anchor_class
      - in the way of the anchor_class
      - preventing a clear view of the anchor_class
\end{lstlisting}

\begin{figure*}
    \centering
    \includegraphics[width=0.8\linewidth]{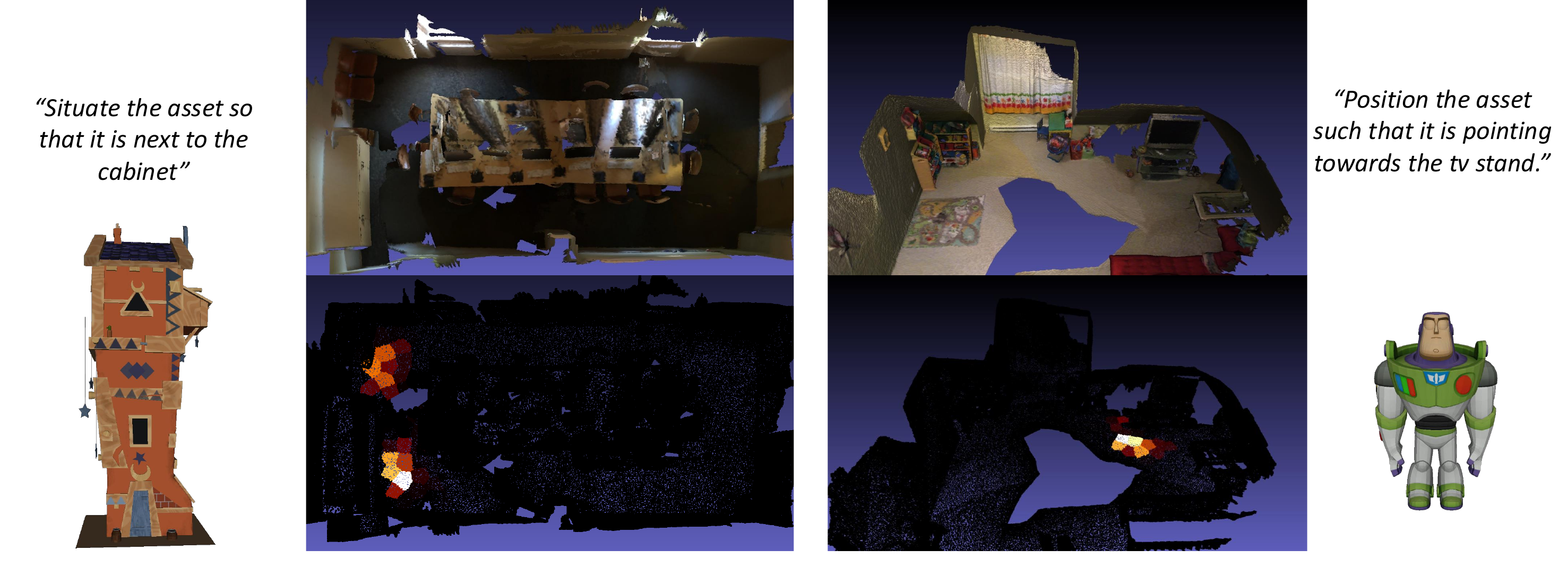}
    \caption{Heatmap visualization of the predicted confidence scores by our model for spatial clusters in two examples from our dataset, across two scenes, given different assets and textual prompts. Warmer colors indicate higher confidence regions for asset placement, with white representing the highest confidence.}
    \label{fig:heatmap}
\end{figure*}
 
\begin{figure*}
    \centering
    \includegraphics[width=0.8\linewidth]{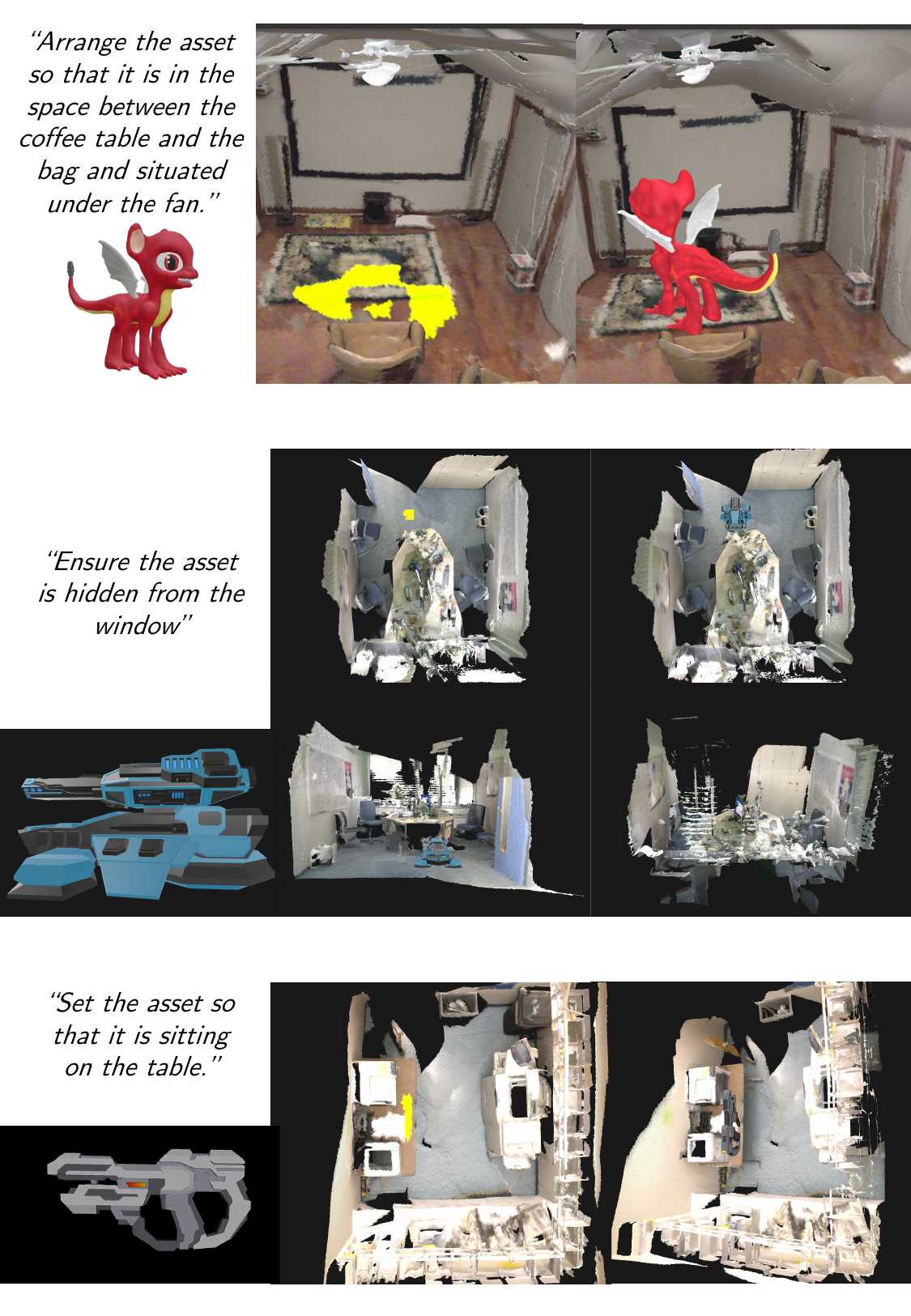}
    \caption{Examples from our proposed dataset illustrating prompts with different constraints, along with the corresponding placement mask and a sample placed asset.}
    \label{fig:dataset_example_2}
\end{figure*}

\begin{figure*}
    \centering
    \includegraphics[width=0.8\linewidth]{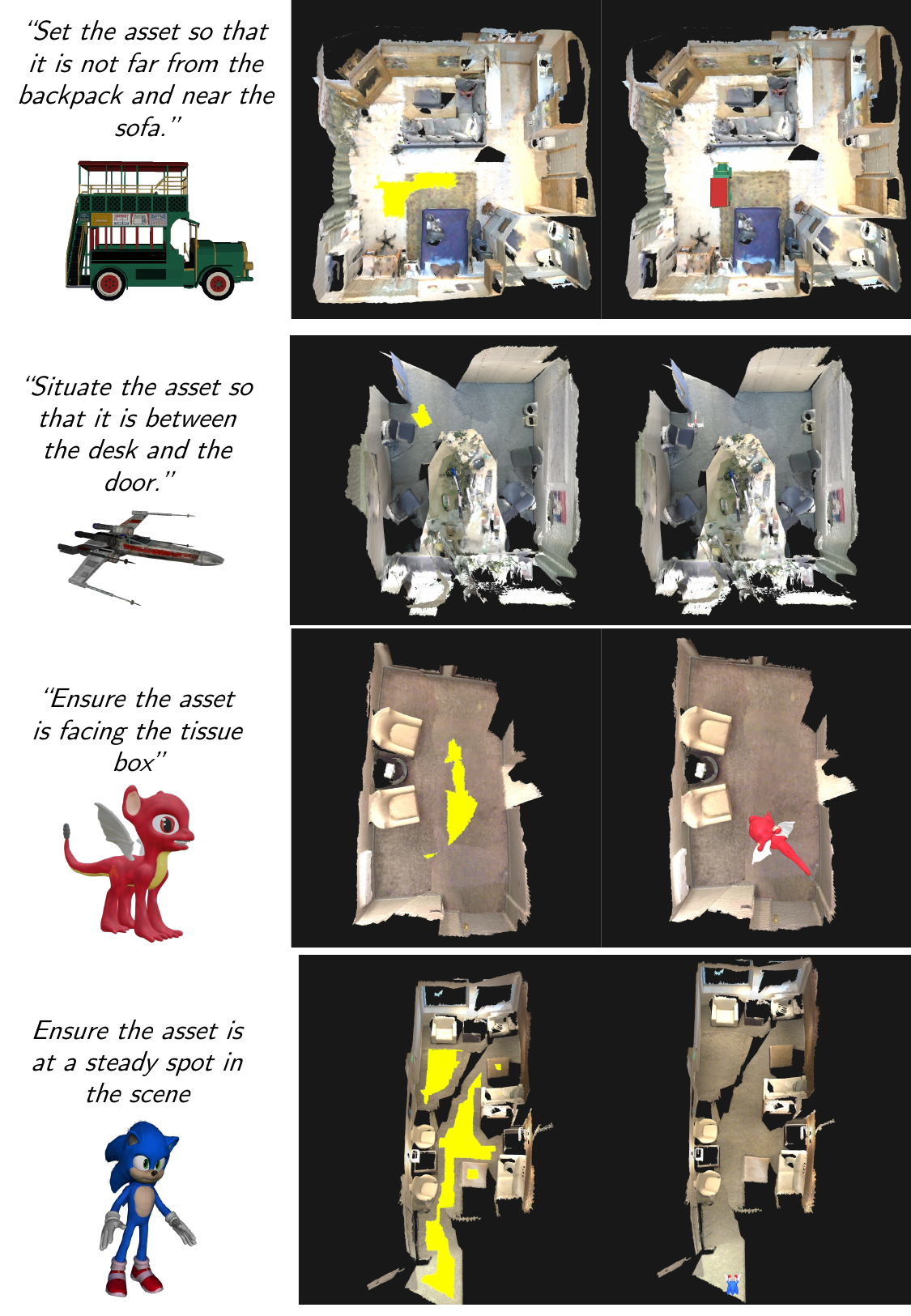}
    \caption{Examples from our proposed dataset illustrating prompts with different constraints, along with the corresponding placement mask and a sample placed asset.}
    \label{fig:dataset_example}
\end{figure*}

\paragraph{Dataset Examples}
In Figures~\ref{fig:dataset_example_2} and ~\ref{fig:dataset_example}, we provide examples from our proposed dataset.

\section{PlaceWizard Implementation Details}
We conduct all our experiments using eight NVIDIA Tesla A100 GPUs, with a training batch size of 28 per single gpu. Following Reason3D, we utilize the AdamW optimizer with parameters $\beta_1 = 0.9$ and $\beta_2 = 0.999$, a weight decay of 0.05, and a linear warm-up strategy for the learning rate during the initial 1000 steps gradually increasing it from $10^{-8}$ to $10^{-4}$ followed by a cosine decay schedule. We train for 50 epochs. We also use a pretrained FlanT5\-XL model, keeping most of its pre-trained weights frozen, except for adapting the weights of the newly added tokens, as similarly done in Reason3D. For spatial pooling, we employ 1024 groups for each ScanNet scan.

\section{Visualization of superpoints}
Figure \ref{fig:superpoint} shows the difference between the superpoints \cite{landrieu2018large} used in Reason3D \cite{huang2024reason3d} and our proposed spatial pooling. While \cite{landrieu2018large} generates large clusters, such as for the floor, our method produces clusters at a finer granularity.

\section{Further Qualitative Results}
In Figure~\ref{fig:heatmap} we show the confidence scores predicted by our model for the spatial clusters in two example scenes from our dataset.

\newpage

\end{document}